\documentclass[conference]{IEEEtran}
\usepackage[letterpaper, left=1.02in, right=1.02in, bottom=1in, top=0.75in]{geometry}
\usepackage{cite}
\usepackage{amsmath,amssymb,amsfonts}

\usepackage[pdftex]{graphicx}
\usepackage{color}

\usepackage{caption,subcaption}
\usepackage{array, booktabs, multirow}
\newcommand{\PreserveBackslash}[1]{\let\temp=\\#1\let\\=\temp}
\usepackage{algorithm}
\usepackage{algpseudocode}

\definecolor{myred}{rgb}{1,0.2,0.2}
\definecolor{myblue}{rgb}{0,0.3,1}
\definecolor{mygreen}{rgb}{0.2,0.7,0}
\definecolor{myorange}{rgb}{1,0.5,0}

\usepackage{textcomp}
\def\BibTeX{{\rm B\kern-.05em{\sc i\kern-.025em b}\kern-.08em
    T\kern-.1667em\lower.7ex\hbox{E}\kern-.125emX}}
\begin{document}

\title{{Drone Positioning for Visible Light Communication with Drone-Mounted LED and Camera}
}

\author{
\IEEEauthorblockN{1\textsuperscript{st} Yukito Onodera}
\IEEEauthorblockN{4\textsuperscript{th} Yu Nakayama}
\IEEEauthorblockA{
\textit{Department of Computer and Information Sciences} \\
\textit{Tokyo University of Agriculture and Technology}\\
Tokyo, Japan \\
\{y.onodera, yu.nakayama\}@ieee.org}
\and
\IEEEauthorblockN{2\textsuperscript{nd} Hiroki Takano}
\IEEEauthorblockN{3\textsuperscript{rd} Daisuke Hisano}
\IEEEauthorblockA{
\textit{Graduate School of Engineering}\\
\textit{Osaka University}\\
Osaka, Japan\\
takano@pn.comm.eng.osaka-u.ac.jp\\
hisano@comm.eng.osaka-u.ac.jp}
}

\maketitle

\begin{abstract}
The world is often stricken by catastrophic disasters.
On-demand drone-mounted visible light communication (VLC) networks are suitable for monitoring disaster-stricken areas for leveraging disaster-response operations.
The concept of an image sensor-based VLC has also attracted attention in the recent past for establishing stable links using unstably moving drones.
However, existing works did not sufficiently consider the one-to-many image sensor-based VLC system.
Thus, this paper proposes the concept of a one-to-many image sensor-based VLC between a camera and multiple drone-mounted LED lights with a drone-positioning algorithm to avoid interference among VLC links.
Multiple drones are deployed on-demand in a disaster-stricken area to monitor the ground and continuously send image data to a camera with image sensor-based visible light communication (VLC) links.
The proposed idea is demonstrated with the proof-of-concept (PoC) implemented with drones that are equipped with LED panels and a 4K camera.
As a result, we confirmed the feasibility of the proposed system.
\end{abstract}


%
%
\section{Introduction}
The world is often stricken by catastrophic disasters including earthquakes, hurricanes, and tsunamis~\cite{Gabe2005harricane, nojima2016earthquake}.
Telecommunication services become unavailable in disaster-stricken areas because of destruction of facilities such as fiber cables or disruption of energy supply.
It has been a significant issue to quickly recover telecommunication networks in such areas for disaster-response and life-saving operations.
Although recovery schemes using surviving facilities by establishing wireless bypass routes were proposed~\cite{nakayama2016wired, nakayama2017wired, nakayama2018recovery}, it is difficult to employ such approaches if massive power outage occurs.
A promising solution for post-disaster monitoring is the use of drone-mounted wireless networks because many drones can be flexibly deployed on-demand in disaster-stricken areas.
On-demand drone-mounted networks are suitable for grasping the current situation in the target area for leveraging disaster-response operations such as finding injured people.

The concept of drone empowered wireless networks has been a hot research topic to deploy public safety networks~\cite{restas2015drone, rabta2018drone}.
A stochastic geometry based design of flying cellular networks was proposed in \cite{hayajneh2016drone} for post-disaster situations.
The effectiveness of flying networks was demonstrated in this work; the ground coverage can be enhanced by optimally selecting the number of drones and their corresponding altitudes.
Also, drone-mounted LTE femtocell base stations were investigated in \cite{deruyck2016emergency} to enhance saturated existing ground infrastructure.
They presented initial results to show the validity of the flying base stations although a large number of drones are required to cover all users in a city.
The concept of VLC on drones was studied in \cite{yang2019power} to simultaneously provide flexible communication and illumination.
The locations of nodes and cell associations were optimized to minimize the power consumption under illumination and communication constraints.
However, the existing drone-mounted wireless networks did not focus on constructing ad hoc networks by many drones for post-disaster monitoring.

To address this problem, this paper proposes a concept and preliminary results of a visible light ad hoc network using multiple drones and an image sensor.
The idea behind the proposed approach is to utilize on-board LED lights for both lighting and communication in a blackout area.
Multiple drones are deployed on-demand in a disaster-stricken area to monitor the ground and continuously send image data to a camera with image sensor-based visible light communication (VLC) links.
This paper also proposes a positioning algorithm for multiple drones to avoid interference among VLC links.
This is because the camera receives optical signals from multiple drone-mounted LEDs, and thus the drones must move to avoid overlap satisfying the requirements for filming, i.e. the recognizable range of drones is determined by conditions including focal length of the camera.
The proposed idea is demonstrated with the proof-of-concept (PoC) implemented with drones that are equipped with LED panels and a 4K camera.

The rest of this paper is organized as follows.
Section \ref{sec:rltd} introduces related work on VLC and the contribution of this work.
Section \ref{sec:pro} describes the proposed scheme.
Section \ref{sec:alg} presents the proposed positioning algorithm and simulation results.
Section \ref{sec:exp} describes the experimental results with the implemented PoC.
Finally, the conclusion is provided in Section \ref{sec:cncl}.

%
%
\section{Related work} \label{sec:rltd}
Free space optical (FSO) communication using drones has been intensely studied in the recent past.
For rapid event response and flexible deployment, an edge-computing-empowered radio access network architecture with drones was proposed in \cite{dong2018edge} where the fronthaul and backhaul links were established with FSO communication between drones and ground nodes.
The turbulence induced signal fades in ad hoc mesh FSO network was experimentally measured in \cite{perez2014experimental} to obtain knowledge on the effect of the attenuation and phase distortion of atmospheric channel.
An FSO based drone assisted mobile access networks was investigated in \cite{wu2019fso} for disaster-response purpose.
The deployment of drones and mobile user association were jointly optimized to maximize the number of served users in a disaster-stricken area.
However, the significant difficulty in FSO-based communication is configuring the light axis between the transceiver and the receiver.
If the light axis is mismatched, the received optical power decreases drastically.
This characteristic makes FSO suitable for communication between fixed ground nodes.
Since most of the drones unstably move in the air due to wind, it is difficult to dynamically adjust the light axis.

To address this issue, the concept of a camera system receiving the optical signal has attracted much attention.
In \cite{takai2014optical}, a VLC-based vehicle-to-vehicle (V2V) communication system was developed, where LED transmitters and video cameras are mounted on vehicles.
As regards VLC between drone-mounted LED lights and a camera, PoC test results using a drone and a ground base station were reported in \cite{chhaglani2020evaluating}.
However, the major limitations of this work are 1) to measure the signal quality between one drone and one camera, and 2) not to recognize the position of the drone in an image.
Therefore, the contribution of this paper is to propose the concept of a one-to-many image sensor-based VLC between a camera and multiple drone-mounted LED lights with a drone-positioning algorithm to avoid interference among VLC links.
Also, we present the preliminary results for the proposed scheme.

%
%
\section{Proposed scheme} \label{sec:pro}
\subsection{Concept} \label{sec:pro_cncpt}
\begin{figure}[!t]
\centering
	\includegraphics[width=3.0in]{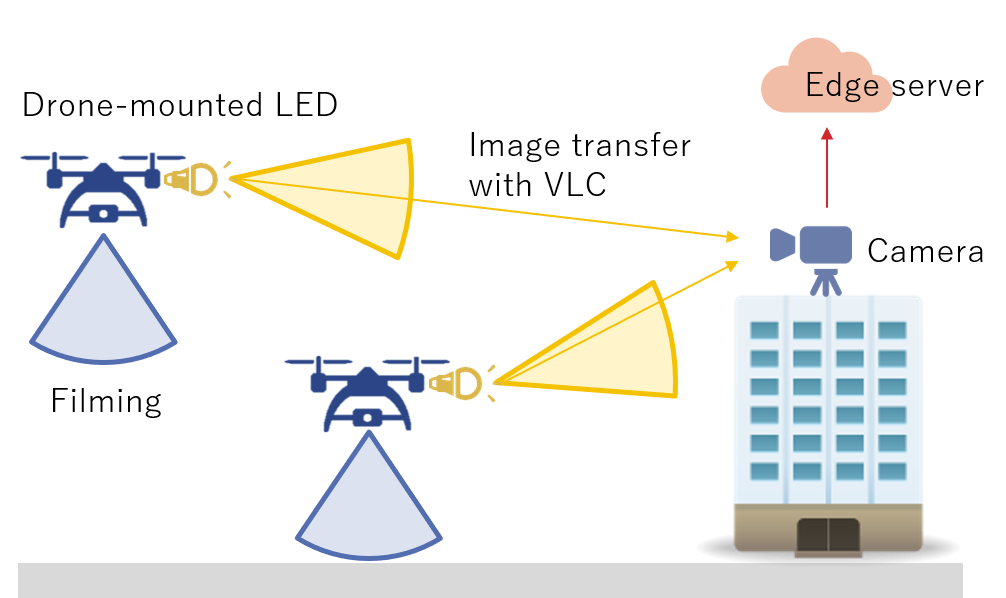}
	\caption{Concept of VLC between drone-mounted LED and camera.}
	\label{fig:concept}
\end{figure}

The conceptual system architecture of the VLC between drone-mounted LED lights and a camera is depicted in Fig.~\ref{fig:concept}.
Multiple drones are deployed in a disaster-stricken area to compose an ad hoc network for post-disaster monitoring.
Each drone films the ground with an on-board camera.
The drones transfer the recorded image to a ground camera with VLC using on-board LED lights.
The camera films the drones and the LED lights mounted on them and sends the data to an edge server.
The edge server detects the drones from received images and demodulates the signals of LED lights.
Note that the edge server can demodulate the signals of LED lights only after the drone-detection.

The goal of the proposed drone positioning algorithm is to achieve efficient data transmission with the VLC links.
The trajectorys of drones are determined to avoid overlap while the requirements for filming are satisfied and the recognizable range of drones is ensured considering external conditions.
The drone positioning model for VLC is defined in the following, and the positioning algorithm is proposed based on the model.
The proposed scheme enables real-time monitoring of the disaster-stricken area for leveraging recovery operations.
The use of visible light is suitable for post-disaster monitoring because of the wide coverage of target area and the multi-purpose use.

\subsection{Image processing sequence} \label{sec:pro_prcs}
\begin{figure}[!t]
\centering
	\includegraphics[width=2.5in]{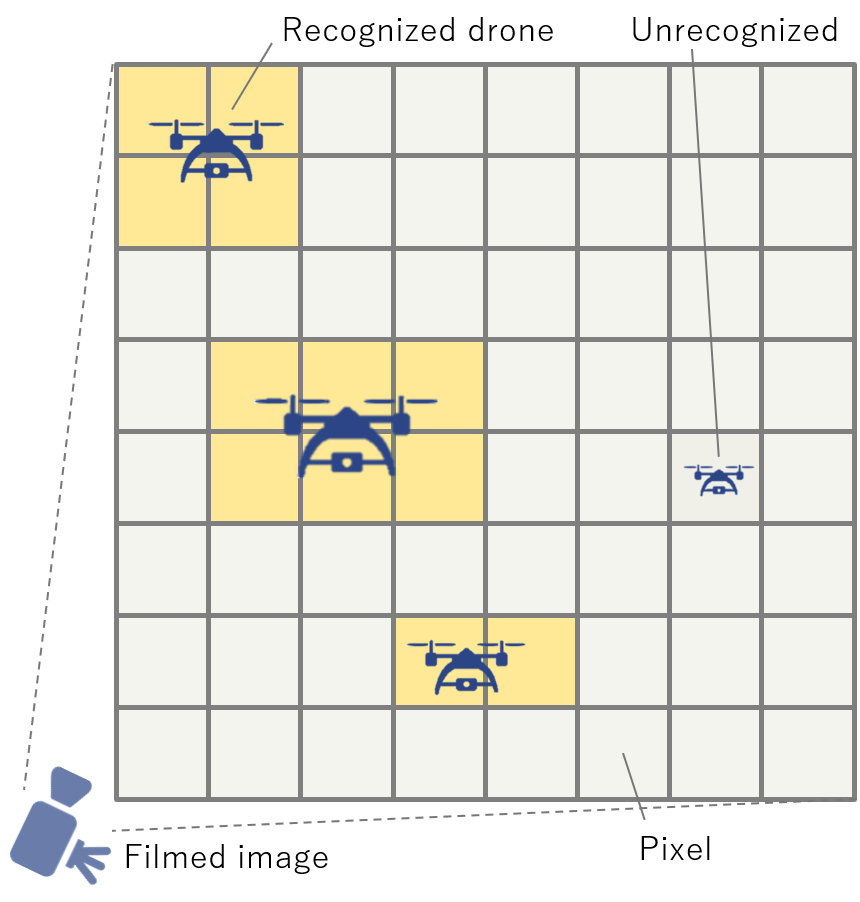}
	\caption{Image filmed by camera.}
	\label{fig:pixel}
\end{figure}

This section introduces the sequence of the image processing at the edge server.
Fig.~\ref{fig:pixel} depicts an example of an image filmed by the camera.
The edge server first detects the drones with DNNs.
Since multiple drones are captured in the image, it then crops the image to extract each drone from the original file.
For each cropped image, the signals of drone-mounted LED lights are detected and demodulated.
Note that the demodulation process can be executed only if the drone-detection succeeded, otherwise the edge server cannot distinguish on-board LEDs from other light sources such as street lights.
A drone which is located too far can be too small to be recognized.

\subsection{Model} \label{sec:pro_mdl}
\subsubsection{Variable definition} \label{sec:pro_mdl_var}
The variables used in the proposed model are summarized in Table~\ref{tbl:variables}.
The detail of each variable is explained in the following.
In the proposed model, we assumed that a drone is approximated as a sphere without loss of generality.

\let\PBS=\PreserveBackslash
\begin{table}[!t]
	\renewcommand{\arraystretch}{1.2}
	\caption{Variables}
	\label{tbl:variables}
	\centering
	\begin{tabular}[t]{>{\PBS\centering\hspace{0pt}}p{0.75in} >{\PBS\centering\hspace{0pt}}p{2.0in}} \toprule
		Variable & Definition \\ \hline
		$\mathcal{I}$ & Set of drones \\
		$i, j$ & Drone identifier in $\mathcal{I}$ \\
		$(x_{i}, y_{i}, z_{i})$ & Position of drone in $(x, y, z)$ space \\
		$r_{i}$ & Distance between $i$th drone and camera \\

    	$d_r$ & Radius of drone \\
		$M$ & Optical zoom magnification of camera \\
		$f$ & Focal length of camera \\

		\bottomrule
	\end{tabular}
\end{table}
\renewcommand{\arraystretch}{1}

\subsubsection{Coordinates of drones} \label{sec:pro_mdl_drn}
Let $\mathcal{I}$ denote the set of drones, and $i$ and $j$ denote identifiers for them.
Assuming that each drone measures its current position using sensors such as Global Positioning System (GPS) sensors.
The origin represents the position and the y-axis denotes the direction of the camera, respectively.
The distance $r_{i}$ is calculated from $(x_{i}, y_{i}, z_{i})$.


\subsubsection{Detectable range} \label{sec:pro_mdl_rng}
Let us define the optical zoom magnification of the camera as $M$.
Let $d_{w}$ and $d_{h}$ denote the width and the height of a drone, assuming that the size of all the drones in $\mathcal{I}$ is the same.
With $M$, we have
%

\begin{equation}
M = \frac{e_{r}}{d_{r}},
\label{eq:cmr_M}
\end{equation}

where $e_{w}$ and $e_{h}$ denote the width and height of the imaged drone in the camera, respectively.

\begin{figure}[!t]
\centering
	\includegraphics[width=3.0in]{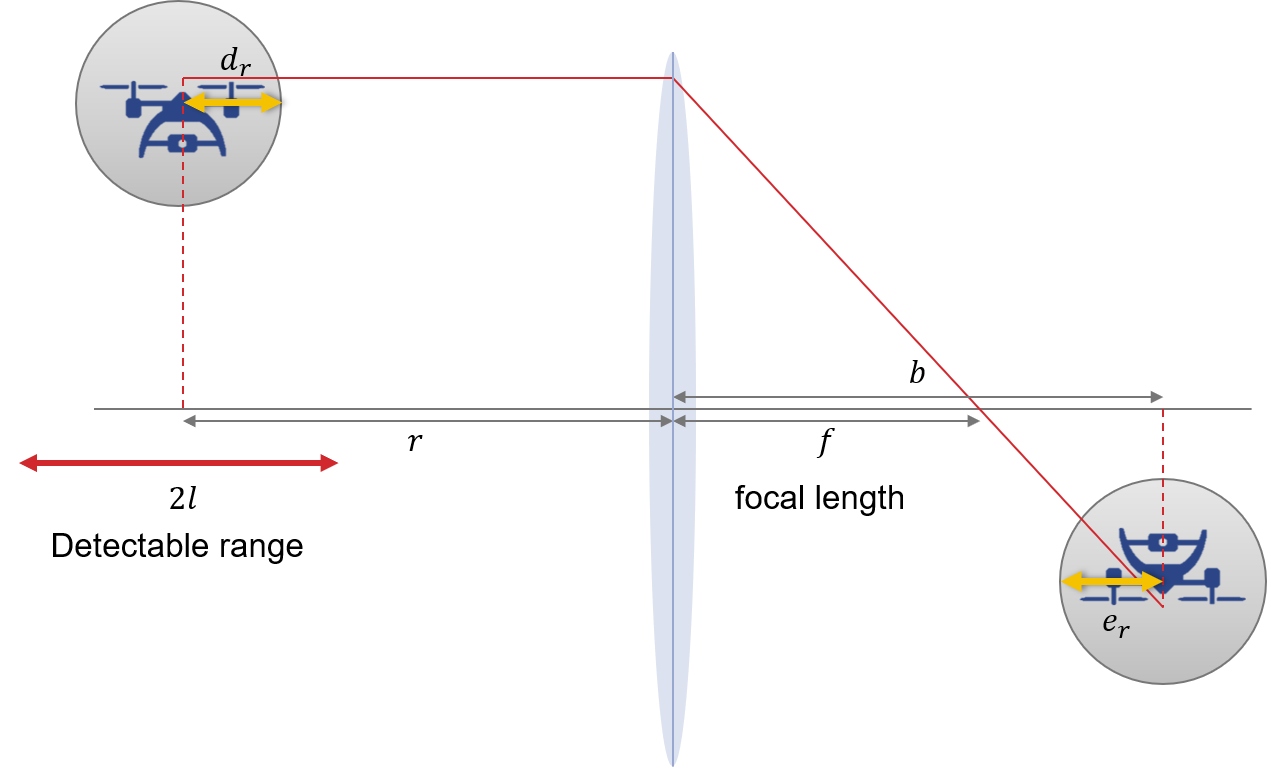}
	\caption{Size of imaged drone.}
	\label{fig:lens}
\end{figure}

Here, we have the general equation for a lens:
\begin{equation}
\frac{1}{r} + \frac{1}{b} = \frac{1}{f},
\label{eq:cmr_lens}
\end{equation}
where $r$ is the distance between the drone and the lens, $b$ is the distance between the imaging point and the lens, and $f$ is the focal length of the camera.
Fig.~\ref{fig:lens} shows the relationship between them.
Using $M = \frac{b}{r}$, \eqref{eq:cmr_lens} is transformed as
\begin{equation}
r = \frac{M + 1}{M} f.
\label{eq:cmr_r}
\end{equation}
The ranges of $M$ and $f$ are determined by the specification of the camera.
With \eqref{eq:cmr_r}, the distance between the drone and the imaging point is formulated as
\begin{align}
L &= r + b \\
  &= (M + 1) r \\
  &= \frac{(M + 1)^{2}}{M} f.
\label{eq:cmr_l}
\end{align}
Thus, the range of focusable distance is determined by the parameters $M$ and $f$.
Note that a drone located within a certain range from the distance $L$ can also be recognized.
This detectable range is defined as $2l$, i.e. a drone located at $(L - b) - l \leq r \leq (L - b) + l$ can be detected.

%


\subsubsection{Overlap of drones} \label{sec:pro_mdl_ovl}
\begin{figure}[!t]
\centering
	\includegraphics[width=3.0in]{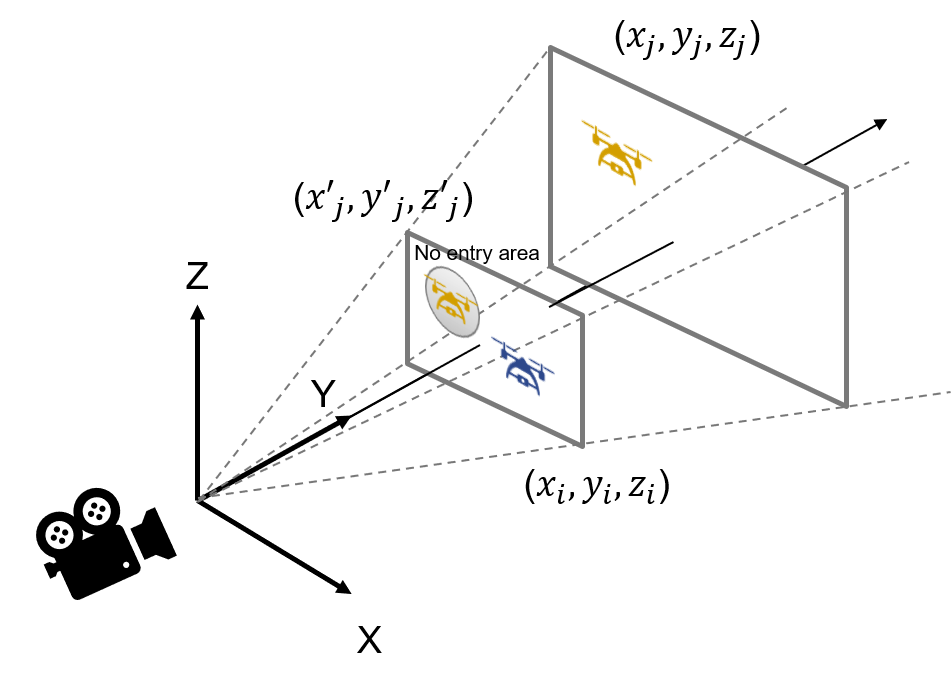}
	\caption{No-entry area of drone.}
	\label{fig:noentry}
\end{figure}

Here we model the overlap of drones in the filmed image, because overlapping drones cannot be separately recognized.
The following model is explained in the x-z plane, because the y-axis represents the direction of the camera.
We define a no-entry area for avoiding overlap of drones.
Fig.~\ref{fig:noentry} depicts the no-entry area for $i$th drone generated by $j$th drone.
The coordinates of $j$th drone which is projected in the $i$th drone's x-z plane is formulated as
\begin{align}
(x'_{j}, y'_{j}, z'_{j}) = \left(\frac{y_{i} x_{j}}{y_{j}}, y_{i}, \frac{y_{i} z_{j}}{y_{j}} \right).
\label{eq:noentry_p}
\end{align}
The radius of the projected $j$th drone is computed as
\begin{equation}
_{i}d'_{j} = \left| \frac{y_{i} (r_{j} + x_{j})}{y_{j}} - x'_{j} \right|.
\end{equation}

The distance between $i$th drone and $j$th drone in the $i$th drone's x-z plane is calculated as
\begin{equation}
\delta_{ij} = \sqrt{(x'_{j} - x_{i})^{2} + (y'_{j} - y_{i})^{2}}.
\end{equation}
Thus, the constraint that there is no overlap of $i$th and $j$th drones in the filmed image is described as
\begin{equation}
\delta_{ij} > d_{r} + _{i}d'_{j}.
\end{equation}

\subsubsection{Constraints for drone positioning} \label{sec:pro_cnst}
Based on the above model, the constraints for drone positioning are summarized as the following.

\paragraph{Detectable range} \label{sec:pro_cnst_rng}
All the drones in $\mathcal{I}$ are within the detectable range:
\begin{equation}
(L - b) - l \leq r_{i} \leq (L - b) + l \quad \forall i \in \mathcal{I}.
\end{equation}

\paragraph{No overlap} \label{sec:pro_cnst_ovl}
All the drones in $\mathcal{I}$ are not mutually overlapped in the image:
\begin{equation}
\delta_{ij} > d_{r} + _{i}d'_{j} \quad \forall i,j \in \mathcal{I}.
\end{equation}

%
%
\section{Drone positioning algorithm} \label{sec:alg}
This section describes the proposed drone positioning algorithm and evaluation results with computer simulations.

\subsection{Assumption} \label{sec:alg_ass}
This paper makes the following assumption on the movement of drones.
\begin{itemize}
\item The destination of each drone is given; each drone moves to its own destination for the monitoring purpose.
\item The moving distance is preferred to be short considering the energy consumption.
\item Each drone has the information on the destination coordinates and the current coordinates of itself and other drones.
\end{itemize}

\subsection{Algorithm} \label{sec:alg_alg}
\begin{figure}[!t]
\centering
	\includegraphics[width=3.0in]{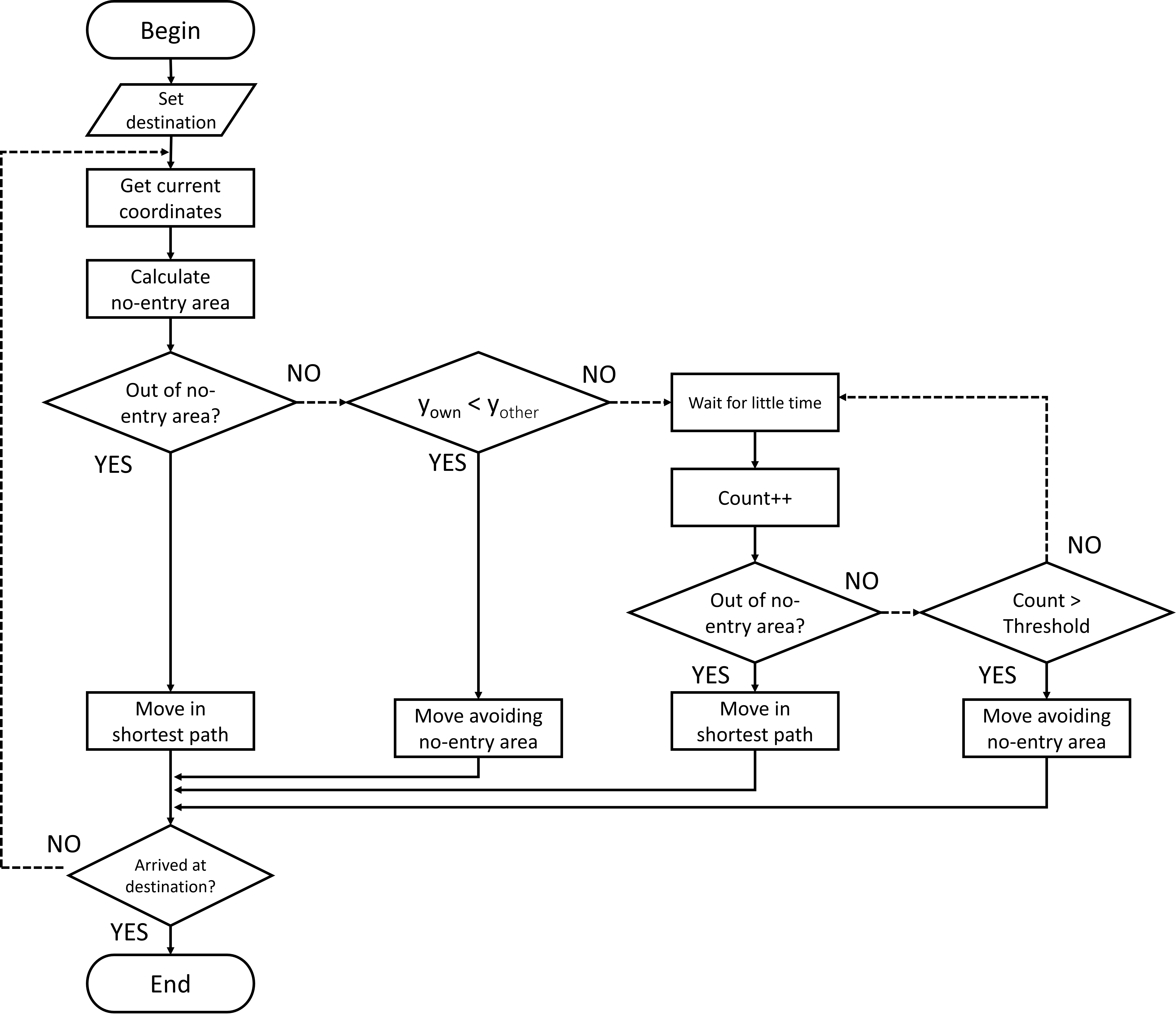}
	\caption{Flowchart of the proposed algorithm.}
	\label{fig:flowchart}
\end{figure}

The goal of the proposed algorithm is to reduce the moving distance of drones satisfying the constraints formulated in section~\ref{sec:pro_cnst}.
The flowchart of the proposed algorithm is shown in Fig.~\ref{fig:flowchart}.
The destination coordinates are given for the drones.
Each drone gets its current coordinates using GNSS and reports to others via wireless communication.
Then, the no-entry area is updated with the constraints described in section \ref{sec:pro_cnst} based on their current coordinates.
Basically, a drone moves in the shortest path to its destination.
When a drone is entering the no-entry area, it identifies the opponent to overlap.
Let $y_{own}$ and $y_{other}$ denote the y-coordinate of this drone and the opponent, respectively.
If $y_{own} < y_{other}$ is satisfied, the drone moves to avoid the no-entry area to reduce the length of bypass route.
Otherwise, the drone stops to wait for the opponent to avoid the no-entry area.
Then, it restarts to move if the no-entry area in the shortest path is eliminated.
If the drone has to wait more than a certain period of time, it starts to move avoiding the no-entry area.
As a result, all the drones can arrive at their destinations without overlap in the filmed images.

\subsection{Simulation results} \label{sec:alg_rslt}
The performance of the proposed algorithm was confirmed with a self-developed simulator written in Python3.
We deploy $8$ drones with $d_{r} = 0.12$ m in a $6$ m square area.
They moved with a speed of $1$ m/s from start positions to destinations, which were both randomly determined.
The flying heights ranged from $3$ m to $7$ m.
The simulations were iterated for $10$ times.

\begin{figure}[!t]
\centering
	\includegraphics[width=2.5in]{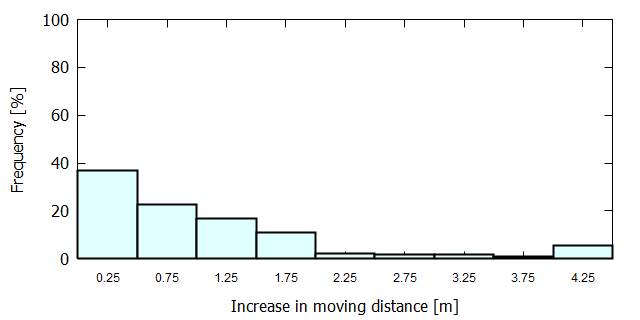}
	\caption{Increase in moving distance.}
	\label{fig:increase}
\end{figure}

\begin{figure}[!t]
\centering
	\includegraphics[width=2.5in]{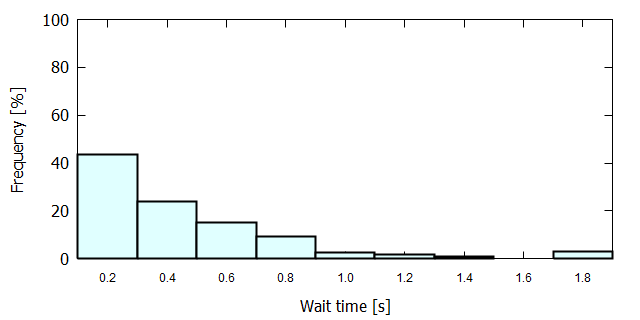}
	\caption{Wait time.}
	\label{fig:waittime}
\end{figure}

Fig.~\ref{fig:increase} shows the distribution of increase in the moving distance of drones.
Note that only the drones that moved to avoid no-entry areas were counted.
The increased distance was under one meter in about $50$\% of cases.
Fig.~\ref{fig:waittime} summarizes the wait time of drones.
The increase in the wait time was within one second in most of cases.
From these results, it was confirmed that drones efficiently move to avoid overlap with the proposed algorithm.

\begin{figure}[!t]
\centering
	\includegraphics[width=3.0in]{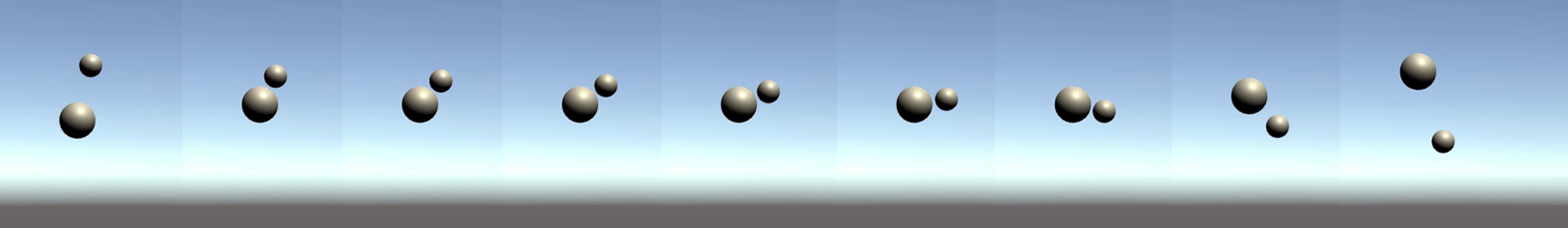}
	\caption{Example case of two drones.}
	\label{fig:motion}
\end{figure}

Furthermore, we demonstrate an example of avoiding movement.
The proposed algorithm was implemented with Unity 2019.4.12.
Fig.~\ref{fig:motion} shows an example case of two drones where each sphere represents a drone.
The drones moved straight to get closer.
Then, a drone avoided the no-entry area, while the other stopped to wait.
Finally, the stopped drone restarted to move.
As a consequence, overlap in the filmed image was successfully avoided.


%

%
%
\section{Experimental results} \label{sec:exp}
This section provides the experimental results of drone recognition.
We obtained $8000$ datasets of drone-mounted LED lights at different environments.
Then, the recognition accuracy was evaluated with a convolutional neural network (CNN).

\subsection{Experimental condition} \label{sec:exp_cnd}
\begin{figure}[!t]
\centering
	\includegraphics[width=2.5in]{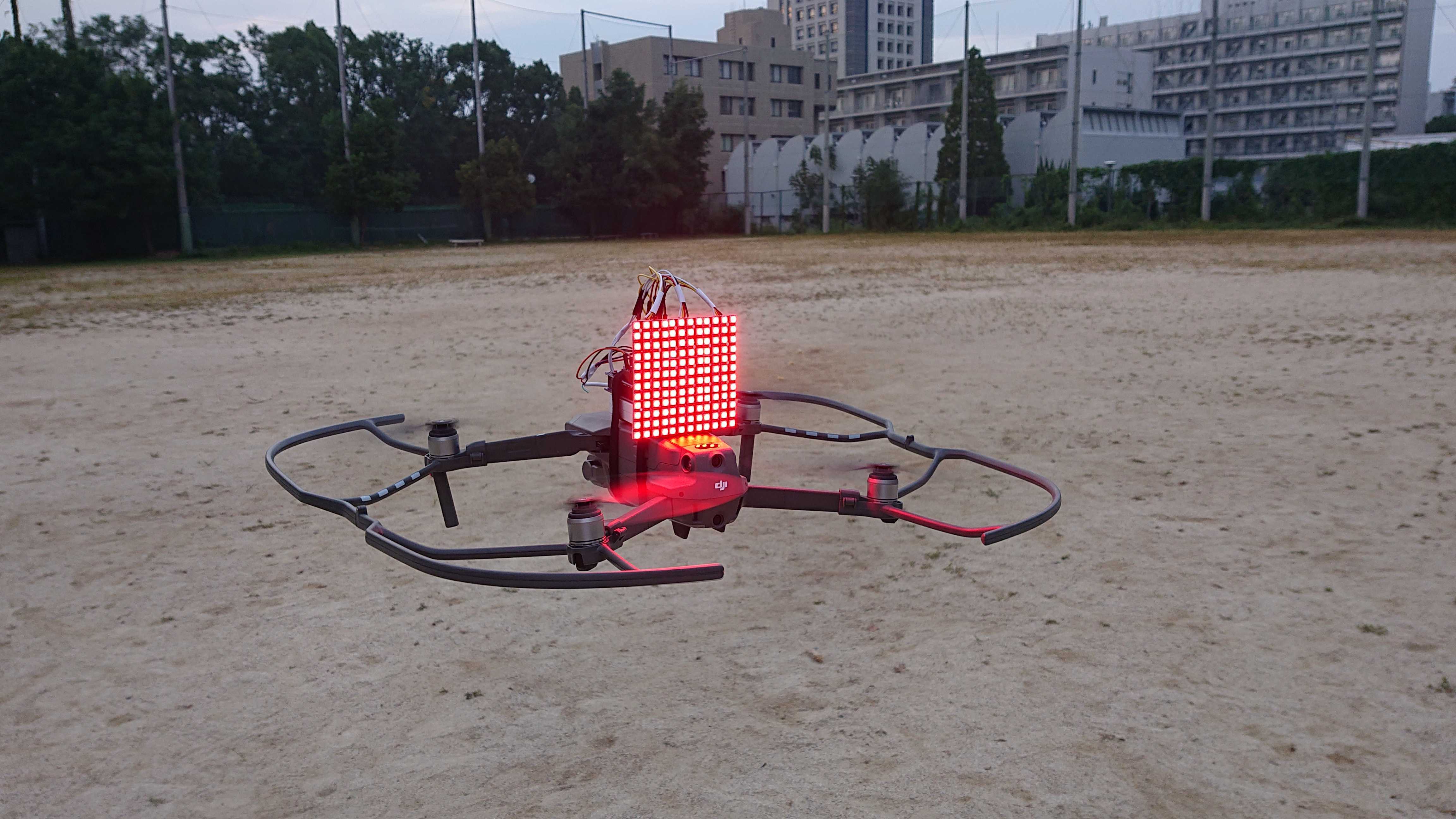}
	\caption{Experimental device.}
	\label{fig:drone}
\end{figure}

Here we explain the experimental conditions.
We employed two Mavic $2$ Pro drones, launched by DaJiang Innovations Science and Technology (DJI) Co., Ltd.
The size of the drone was approximated as $d_{r} = 12$ cm.
As regards LED lights, we used WS$2812$B serial LED panels with $16$ LED lights placed in square, which were produced by World semi Co., Limited,
Nine serial LED panels were mounted on each drone as the transmitter.
An Arduino UNO micro controller was connected to the serial LED panels to control them.
The experimental device is shown in Fig.~\ref{fig:drone}.
The brightness of the serial LEDs was set to the maximum value to ensure sufficient transmission distance.
As regards the receiver side, a Sony Xperia XZ $2$ Premium camera was employed.
The resolution was $1080 \times 1920$, the pixel count was $2$ mega pixels, and the frame rate was $60$ fps.
The height of the camera was set to $1.0$ m.
The focal length was $f = 25$ mm, and the zoom magnification was set to $M = 1$.

The transmitter mounted on a drone sent continuous light to be filmed by the camera.
The recorded video was sent to an edge server so that it was divided into a series of static pictures.
We employed YOLOv3, which is a famous machine learning model using a CNN, to recognize the transmitter from the received pictures at the edge server.

To ensure the detection accuracy of drone-mounted LED lights, we collected $8000$ pictures as the training data and marked the positions of the drones.
In this process, the resolution of the pictures was resized to $360 \times 640$.
Also, the datasets were obtained by changing the distance between the camera and the drones to improve the recognition accuracy under different conditions.
This is because the number of pixels representing the drone decreases in accordance with the increase in the distance.

\subsection{Results} \label{sec:exp_rslt}
The results for recognition accuracy is summarized in Table~\ref{tbl:accuracy1}.
Each value is the mean of the recognition rates calculated with $100$ test images for each distance.
It was confirmed from this result that drone-mounted LED lights can be accurately and stably recognized regardless of the distance between the camera and the drone.
Also, we confirmed the feasibility of demodulation of data signals from the transmitter.

\begin{figure}[!t]
\centering
	\includegraphics[width=2.5in]{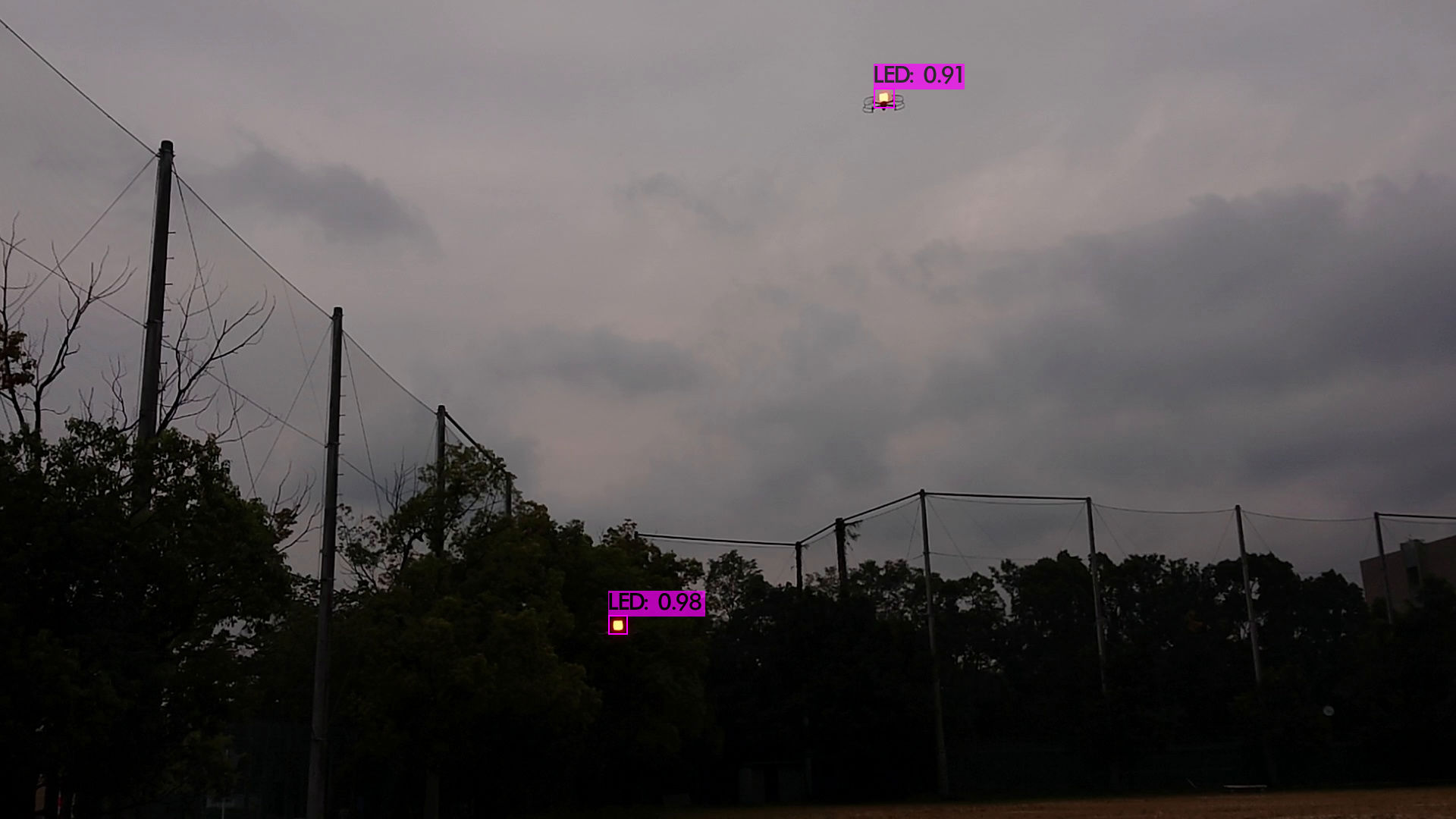}
	\caption{Example recognition result of multiple drones.}
	\label{fig:result_image}
\end{figure}

The validity of the proposed positioning algorithm was confirmed with two drones.
Fig.~\ref{fig:result_image} depicts an example recognition result of two drones in a picture.
In this example, the distances between the camera to each drone were both $20$ m.
From the constraints formulated in section~\ref{sec:pro_cnst}, the no-entry area was computed to avoid overlap.
Since these constraints were satisfied in Fig.~\ref{fig:result_image}, two drones were separately recognized.
Thus, the feasibility of the proposed model and algorithm was confirmed through this result.

\let\PBS=\PreserveBackslash
\begin{table}[!t]
	\renewcommand{\arraystretch}{1.2}
	\caption{Recognition rate}
	\label{tbl:accuracy1}
	\centering
	\begin{tabular}[t]{>{\PBS\centering\hspace{0pt}}p{1.0in} >{\PBS\centering\hspace{0pt}}p{1.0in}} \toprule
		distance [m] & accuracy [\%] \\ \hline
		$10$ & $93.2$ \\
		$20$ & $91.8$ \\
		$30$ & $98.3$ \\
		$40$ & $95.6$ \\
		$50$ & $99.6$ \\
		$60$ & $97.8$ \\
		$70$ & $99.3$ \\
		$80$ & $97.2$ \\
		\bottomrule
	\end{tabular}
\end{table}
\renewcommand{\arraystretch}{1}



\subsection{Discussion} \label{sec:exp_dscs}
The color of LED lights was set to red in the experiment as shown in Fig.~\ref{fig:drone}.
This determination was the result of consideration on the trees and sky in the background.
Since the color which is easy to identify depends on the outer environments, the robust color setting should be investigated.
Also, experiments using more drones is further study.

%
%
\section{Conclusion} \label{sec:cncl}
This paper proposed the concept of a one-to-many image sensor-based VLC between a camera and multiple drone-mounted LED lights.
The proposed idea enables the one-to-many VLC system using unstably moving drones.
With the proposed scheme, multiple drones are deployed on-demand in a disaster-stricken area to monitor the ground and continuously send image data to a ground camera via VLC links.
The on-board LED lights can be utilized for both lighting and communication in a blackout area.
In this paper we also presented a drone-positioning algorithm to avoid interference among VLC links.
This is because the camera receives optical signals from multiple drones, and thus the drones must move to avoid overlap within the detectable range determined by the size of the drone and the focal length of the camera.
The performance of the proposed algorithm was confirmed with computer simulations.
Furthermore, the feasibility of the proposed system was demonstrated with the PoC implemented with drones equipped with LED panels and a $4$ K camera.
It constitutes the future work to evaluate the performance of the proposed idea with experimental results using many drones.

%
%
\section*{Acknowledgment}
A part of this work This work was supported by JST, ACT-I, Grant Number JPMJPR18UL and GMO Foundation, Japan.

%
%
\bibliographystyle{IEEEtran.bst}
\bibliography{bibliography}

\end{document}